\documentclass[10pt,twocolumn,letterpaper]{article}
\usepackage{cvpr}

\usepackage{graphicx,adjustbox}
\usepackage{overpic}
\usepackage{multirow}
\usepackage[table,xcdraw,dvipsnames]{xcolor}
\usepackage{nicefrac}
\usepackage{comment}
\usepackage{xspace}
\usepackage{units}
\usepackage{mdwtab}
\usepackage{url}
\usepackage{array}
\usepackage{rotating}
\usepackage{setspace}
\usepackage{wrapfig}
\usepackage{upgreek}
\usepackage{multirow}
\usepackage{mathdefs}
\usepackage{enumitem}
\usepackage{caption}
\usepackage{subcaption}
\usepackage[pagebackref,breaklinks,colorlinks]{hyperref}

\usepackage{listings}
\usepackage{passive}

\setitemize{noitemsep,topsep=4pt,parsep=4pt,partopsep=0pt,leftmargin=*}
\setenumerate{noitemsep,topsep=4pt,parsep=4pt,partopsep=0pt,leftmargin=*}

\graphicspath{{figures/}}

\title{
\vspace{-0.2in}
Passive Micron-scale Time-of-Flight with Sunlight Interferometry}
\author{\vspace{-.45in} \\
Alankar Kotwal\textsuperscript{1}, Anat Levin\textsuperscript{2}, and Ioannis Gkioulekas\textsuperscript{1}\\
\textsuperscript{1}Carnegie Mellon University, \textsuperscript{2}Technion
}

\def\cvprPaperID{10412}
\def\confName{CVPR}
\def\confYear{2023}

\expandafter\def\expandafter\normalsize\expandafter{%
    \normalsize
    \setlength\abovedisplayskip{3pt}
    \setlength\belowdisplayskip{3pt}
    \setlength\abovedisplayshortskip{3pt}
    \setlength\belowdisplayshortskip{3pt}
}

\begin{document}
    \maketitle
    
    \begin{abstract}\vspace*{-10pt}
We introduce an interferometric technique for passive time-of-flight imaging and depth sensing at micrometer axial resolutions. Our technique uses a full-field Michelson interferometer, modified to use sunlight as the only light source. The large spectral bandwidth of sunlight makes it possible to acquire micrometer-resolution time-resolved scene responses, through a simple axial scanning operation. Additionally, the angular bandwidth of sunlight makes it possible to capture time-of-flight measurements insensitive to indirect illumination effects, such as interreflections and subsurface scattering. We build an experimental prototype that we operate outdoors, under direct sunlight, and in adverse environment conditions such as machine vibrations and vehicle traffic. We use this prototype to demonstrate, for the first time, passive imaging capabilities such as micrometer-scale depth sensing robust to indirect illumination, direct-only imaging, and imaging through diffusers.
\end{abstract}
    \begin{figure}[t]
    \centering
    \includegraphics[width=\linewidth]{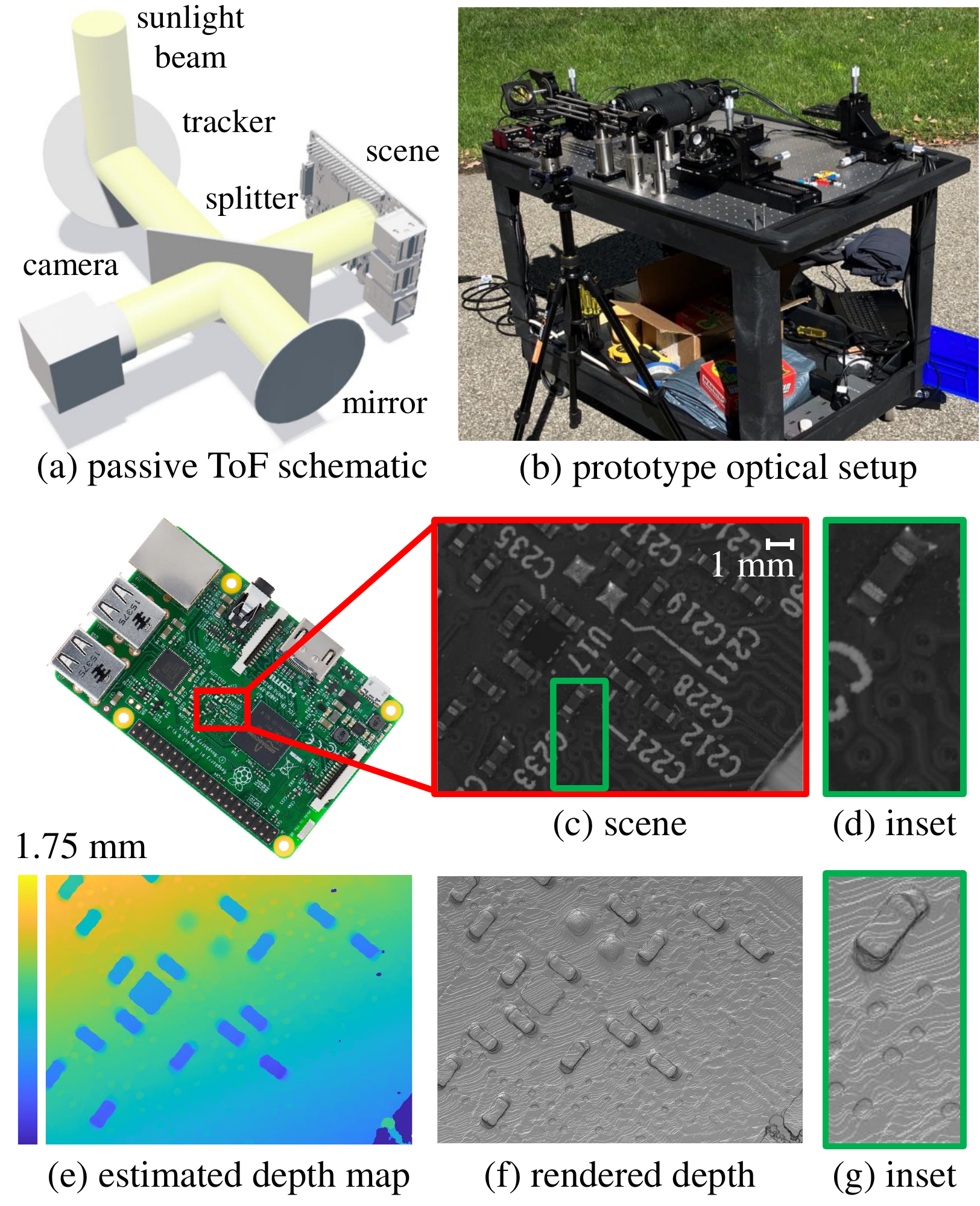}
    \vspace*{-20pt}
    \caption{\textbf{Using sunlight interferometry to passively reconstruct part of a circuit board.} \textbf{(a)} and \textbf{(b)} show a schematic and photograph of the sunlight interferometer we build for passive time-of-flight imaging. We use this system to reconstruct part of a Raspberry Pi circuit board that has multiple resistors, soldering pads, and tracks. \textbf{(c)} and \textbf{(d)} show a picture of the scene as seen through the imaging camera, along with an inset highlighting fine geometric features. \textbf{(e)} shows the estimated depth map, and \textbf{(f)} and \textbf{(g)} the corresponding rendered 3D surface. Our technique reconstructs fine features such as the PCB tracks and through-holes, despite operating outdoors under adverse environment conditions.}
    \label{fig:teaser}
\end{figure}
    \vspace*{-30pt}
\section{Introduction}\label{sec:intro}
\vspace*{-5pt}

The recovery of 3D information for an imaged scene is one of the core problems of optics, imaging, computer vision, and other sciences. In particular, the ability to sense depth at axial resolutions of a few micrometers is of great importance for critical applications areas such as medicine, precision fabrication, material science, and robotics. 

Existing contactless imaging techniques for micron-scale 3D sensing, such as interferometry and microscopy, require active illumination, most commonly from a coherent source. This  makes these techniques impractical for use outdoors, in the presence of strong ambient illumination that overwhelms the active source, or in power-constrained applications. On the other hand, existing passive 3D sensing techniques, such as multi-view stereo, shape from shading, and depth from (de)focus, achieve resolutions of hundreds of micrometers at best, placing them out of scope for applications requiring micron-scale resolutions. 

We change this state of affairs by introducing a completely passive micron-scale 3D sensing technique. Our technique is interferometric, and uses sunlight as the only light source, to capture full-frame time-of-flight measurements at axial resolutions of 5 micrometers. Our technique additionally takes advantage of the spatial incoherence of sunlight, to enable robust 3D sensing in the presence of severe indirect illumination effects (interreflections, subsurface scattering), and even tasks such as imaging and 3D sensing through optically-thin scattering layers. 

To demonstrate these capabilities, we build an experimental prototype that we operate outdoors, under direct sunlight and adverse experimental conditions (wind, machine vibrations, vehicle traffic). This is in stark contrast with previous demonstrations of interferometric sensing, which were only possible under carefully-controlled lab conditions (dark room, vibration-isolated optical tables, no air flow). Even under these adverse conditions, our experiments show that it is possible to perform passive depth scanning, at pixel-level lateral resolution and micrometer axial resolution, for objects that are challenging to scan even with active illumination techniques (translucent, metallic, occluded by thin scatterers). More broadly, our results open the door for the deployment of interferometric techniques in uncontrolled outdoor environments, and for the development of passive computational light transport capabilities such as direct-only imaging, imaging through scattering, and transient imaging. To facilitate future research towards these directions, we provide setup details, reconstruction code, and data in the supplement and project website.%
\footnote{{\scriptsize \url{https://imaging.cs.cmu.edu/sunlight_interferometry}}}

\boldstart{Potential impact.} The ability to perform interferometry outdoors can be useful for applications such as field inspection (e.g., detecting tiny but dangerous defects on airplane surfaces), field robotics (e.g., high-resolution manipulation), and field medicine (e.g., in battlefields, disaster environments, or impoverished regions). In addition to outdoor operation, all of these applications benefit from \emph{passive} operation: Removing the active source allows for lighter, lower-cost, and lower-power depth sensing, all features critical for field operation---often requiring mobility, small form factors, and functionality under limited electricity access. This is particularly important considering that the source is typically the system component consuming the most power, and also one of the costliest and bulkiest.


    \vspace*{-5pt}
\section{Related work}
\vspace*{-5pt}

\boldstart{Passive depth sensing.} A variety of techniques in computer vision measure depth while relying only on external light to illuminate the scene and provide signal to the camera. These techniques use depth cues such as disparity~\cite{Barnard1980disparity,Nalpantidis2008stereo,hartley_zisserman_2004}, focus and defocus~\cite{Hasinoff2006confocal,grossman1987focus,Subbarao1994defocus,hazirbas2018deep,alexander2016focal}, or shading~\cite{horn1970shading,Han2013shading}. Unfortunately, all of these techniques are severely constrained in terms of the reflectance properties of the scene.

\boldstart{Active depth sensing.} To alleviate these limitations, active depth sensing techniques use a controlled light source to inject light into the scene. Examples techniques include photometric stereo~\cite{Woodham1980photometric,zickler2002helmholtz}, structured light~\cite{Scharstein2003structured,gupta2011structured,OToole2016slt,Chen2008structured,o2015homogeneous}, impulse time-of-flight~\cite{kirmani2009looking,Aull2005geiger,Kirmani2014first,gariepy2015single,Gupta2019asingle,Gupta2019bsingle,Heide2018spad,lindell2018single,otoole2017CVPR,Velten2012ultrafast,Niclass2005spadarray,Villa2014cmos}, correlation time-of-flight~\cite{baek2022centimeter,Rochas2003integrated,Lange2001solid,Piatti2013tof,Schwarte1997pmd,heide2013low,Lange2000demodulation}, or combinations thereof \cite{achar2017epipolar}. However, with a few exceptions \cite{achar2017epipolar,o2015homogeneous}, active depth sensing techniques fail under strong ambient lighting conditions that overwhelms the projected illumination. Additionally, their increased power consumption due to the light source can make them impractical for power-constrained applications or field operation.

\boldstart{Interferometric depth sensing.} Most of the techniques listed above are limited to depth resolutions in the order of millimeters (for passive) or hundreds of micrometers (for active). For applications requiring micrometer depth resolution, one option is to use interferometric techniques~\cite{hariharan2003optical,huang1991optical,deGroot2011,johnson2001enhanced,Fercher1985heterodyne,deGroot1992chirped,li2017high,li2018sh,Cheng1984two,Cheng1985multiple,Houairi2009analytical,MeinersHagen2009MultiWavelengthIF,Lu2002step,Yan2015refractive,gkioulekas2015micron,Kotwal2020,Aguirre2015,Wang16cubic}. Unfortunately, all these techniques are active, and extremely sensitive to ambient light. Additionally, the micron-scale sensitivity of interferometric techniques makes them notoriously sensitive to environmental conditions such as vibrations and wind, making outdoor operation seemingly impossible. As a consequence, most interferometric techniques are restricted to controlled lab conditions.

\boldstart{Mitigating indirect illumination.} Active depth sensing techniques typically operate under an assumption of direct-only illumination. Indirect illumination effects, such as interreflections and subsurface scattering, introduces spurious signal into measurements, resulting in inaccurate depth estimation. To mitigate these effects, \emph{computatioanl light transport} techniques focus on \emph{optically} removing indirect illumination by \emph{probing light transport}~\cite{OToole:2012:PCP}. 
Probing techniques use spatial modulation of the illumination and camera, and include schemes such as epipolar imaging~\cite{o2015homogeneous,OToole2016slt}, high-spatial-frequency illumination~\cite{nayar2006fast,reddy2012frequency}, and spatio-temporal code illumination~\cite{o2014temporal,gupta2015phasor,achar2017epipolar}. Similar probing capabilities can be implemented in interferometric systems, by exploiting the spatio-temporal coherence properties of the illumination~\cite{gkioulekas2015micron,Kotwal2020}. Importantly, with the exception of Nayar et al.~\cite{nayar2006fast}, the ability to suppress indirect illumination is currently only available to active techniques.

\boldstart{Passive interferometry.} Passive interferometry has been studied extensively for applications other than outdoor depth sensing. Examples include its use in seismology for subsurface imaging~\cite{schuster2009,Draganov2015PassiveSI,Claerbout1968}, and in ultrasound imaging~\cite{Weaver2001}. More recently, passive interferometry has been used to perform light source localization for occluded imaging applications, in carefully-controlled indoor lab environments~\cite{Davy2013,Badon2015,Badon2016,BogerLombard2019,Batarseh2018}. In radio astronomy, passive interferometry techniques use the Hanbury-Brown-Twiss effect to image astronomically distant light sources~\cite{Jennison1958,Hanbury1954,Brown1956}. Lastly, very-long-baseline interferometry is used for high-resolution imaging of cosmic radio sources, and was used to produce the first-ever image of a black hole~\cite{Bouman2016,Akiyama2019}.
    \vspace*{-5pt}
\section{Theory of sunlight interferometry}\label{sec:background}
\vspace*{-5pt}

We begin by presenting an analysis of interferometric image formation model using sunlight. Our analysis leverages results from previous works on interferometry with incoherent illumination~\cite{gkioulekas2015micron,Kotwal2020}.


\begin{figure}[t]
    \centering
    \includegraphics[width=0.95\linewidth]{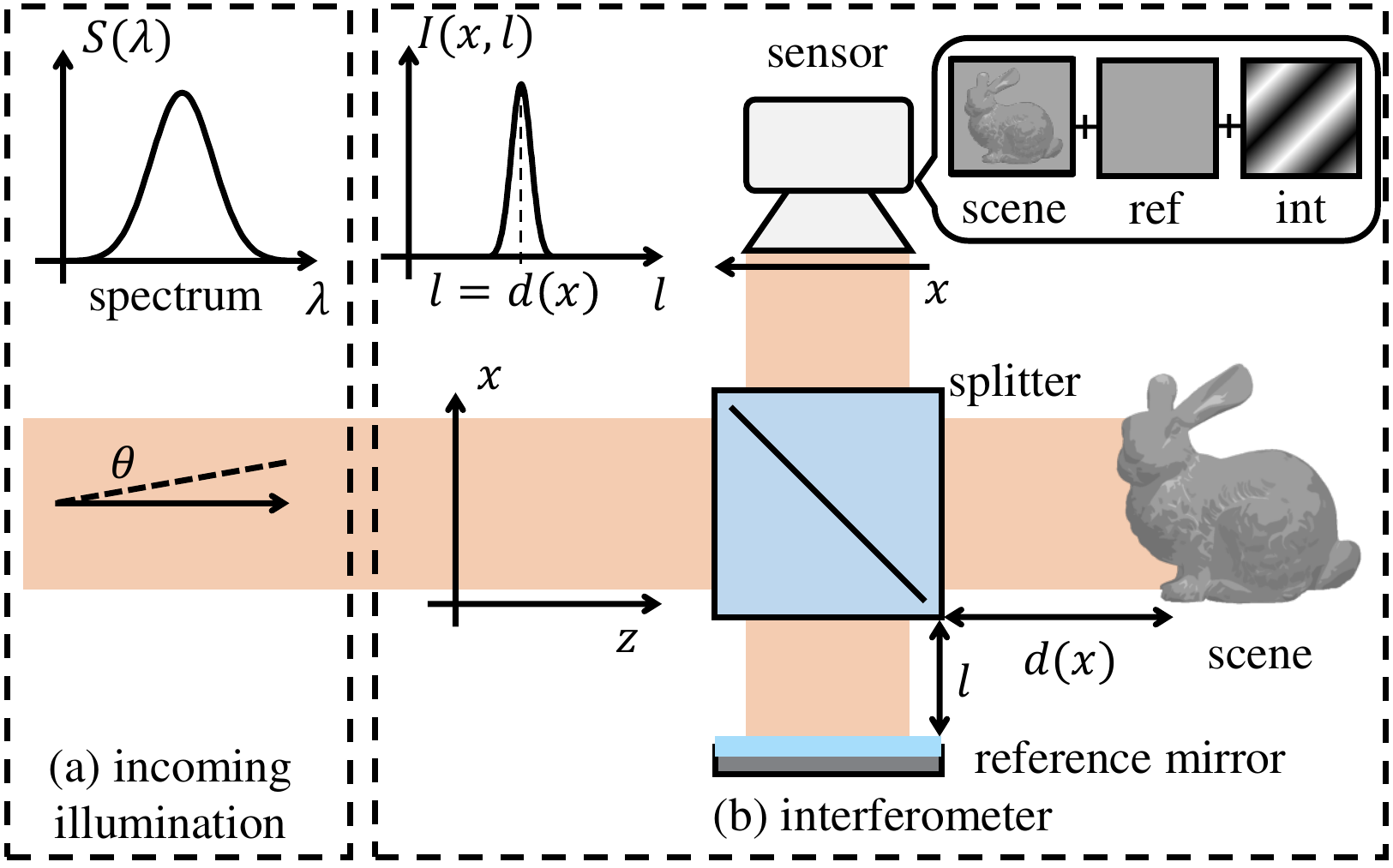}
    \vspace*{-10pt}
    \caption{\textbf{(a)} We use as illumination sunlight, which is both temporally incoherent (large spectral bandwidth) and spatially incoherent (large angular bandwidth). \textbf{(b)} We inject this illumination into a full-field Michelson interferometer, which we use to perform micrometer-scale direct-only time-of-flight imaging.}
    \label{fig:background}
\end{figure}

\boldstart{The Michelson interferometer.} Our optical setup is a version of the classical Michelson interferometer (Figure~\ref{fig:background}(b)). The interferometer receives illumination from direct sunlight, through an optical configuration that reflects sunlight to be parallel with the optical axis (Section~\ref{sec:setup}). A beam splitter splits the input illumination into two beams: one propagates toward the \emph{scene arm}, and the other toward the \emph{reference arm}. The reference arm is a planar mirror mounted on a translation stage that can vary the mirror's distance from the beam splitter. After reflection, the two light beams recombine at the beam splitter and propagate toward a camera with a two-dimensional sensor. 

To simplify notation, we analyze the setup of Figure~\ref{fig:background} in two dimensions---the extension to three dimensions is straightforward. We use an $\paren{x, z}$ coordinate system (Figure~\ref{fig:background}(b)), where $z$ is the optical axis of the interferometer, coinciding with the illumination and sensor optical axes. We denote by $l$ and $d\paren{x}$ the axial distance from the beamsplitter of the reference mirror and scene point that pixel $x$ images, respectively. We assume that the camera focuses at depth $z=d\paren{x}$. As $l$ is a controllable parameter, we denote it explicitly. We denote by $\uref\paren{x,l}$ and $\uscn\paren{x}$ the fields arriving at sensor pixel $x$ after reflection at the reference and scene arms, respectively. Then, the sensor measures,
\begin{equation}
\!\!\!\!I\!\paren{x,l}\! =
	\!\!\underbrace{\abs{\uscn\!\paren{x}}^2\!+\!\abs{\uref\!\paren{x,l}}^2}_{\equiv\, \bi\paren{x,l}}+2\real\!\big\{\!\underbrace{\uscn\!\paren{x}\!\uref^*\!\paren{x,l}}_{\equiv\,\bc\paren{x,l}}\!\big\}.\label{eq:interf-def}
\end{equation}
The first two terms in Equation~\eqref{eq:interf-def} are the intensities the sensor would measure if it were observing each of the two arms separately. We call their sum the \emph{interference-free} image $\bi$.
The third term, which we call \emph{interference}, is the real part of the complex \emph{correlation} $\bc$ between the reflected scene and reference fields. In the rest of this section, we analyze how the correlation changes as a function of the input illumination, reference arm, and scene arm.

\boldstart{Input illumination.} Sunlight, and thus the illumination injected in the interferometer, is temporally and spatially incoherent. \emph{Temporal incoherence} means that the illumination is broadband in spectrum, i.e., comprises multiple \emph{independent} waves propagating with different wavelengths. In practice, as we discuss in Section~\ref{sec:setup}, we use a spectral filter in front of the camera to control temporal incoherence. Rather than the wavelength $\lambda$, we will characterize the illumination spectrum using the wavenumber $\wn\equiv\nicefrac{2\pi}{\lambda}$. Following Goodman~\cite{GoodmanStat}, we model the temporally-incoherent illumination from filtered sunlight by assuming that its power at different wavenumbers $\wn$ follows a Gaussian distribution with mean $\bar{\wn}$ and standard deviation $\Delta_{\wn}$:
\begin{equation}
	\label{eq:temporal}
	\textrm{Power}\paren{\wn}\propto
\exp\paren{-\nicefrac{(\wn-\bar{\wn})^2}{2\Delta_{\wn}^2}}.
\end{equation}
The standard deviation $\Delta_{\wn}$ is the \emph{spectral bandwidth} of the illumination: the smaller its value is, the more monochromatic, and thus temporally coherent, the illumination is.

\emph{Spatial incoherence} means that the illumination comprises multiple planar wavefronts propagating along directions with different angular offsets from the optical axis $z$. This is accurate for sunlight, as the Sun is a far-field area emitter subtending (when observed from the Earth) a small solid angle. Following Goodman~\cite{GoodmanStat}, we model the spatially-incoherent illumination from sunlight by assuming that its power at different angular offsets $\theta$ is uniform within a small range $\Delta_{\theta}$, and zero otherwise:
\begin{equation}
	\label{eq:spatial}
	\textrm{Power}\paren{\theta}\propto \mathbf{1}_{\bracket{-\Delta_{\theta}/2,\Delta_{\theta}/2}}\paren{\theta},
\end{equation}
%
The range $\Delta_{\theta}$ is the \emph{angular bandwidth} of the 
illumination: the smaller its value is, the more collimated, and thus spatially coherent, the illumination is.

\boldstart{The transmission matrix.} To characterize the correlation, we need to consider how the scene arm reflects incident illumination. For this, we use the scene's (continuous) \emph{transmission matrix} $\tc\paren{x,x^\prime}$: This complex scalar function equals the field generated at point $x$ on the in-focus plane $z=d\paren{x}$, given as input an impulse field at point $x^\prime$ on the same plane~\cite{popoff2010measuring,judkewitz2015translation}. The transmission matrix is the complex, wave-based analogue of the scene's \emph{light transport matrix}~\cite{OToole:2012:PCP}. The diagonal elements of the transmission matrix $\tc\paren{x,x}$ equal the wave measured by a confocal system that focuses illumination and sensing at the same point $x$ and depth $d\paren{x}$ in the scene. For a system with coaxial illumination (such as the Michelson interferometer), such elements are dominated by direct illumination and are insensitive to most indirect illumination effects (e.g., interreflections, subsurface scattering)~\cite{OToole:2012:PCP,gkioulekas2015micron}.

\boldstart{Correlation.} We now can characterize the correlation of Equation~\eqref{eq:interf-def}. Given illumination with spectral bandwidth $\Delta_{\wn}$ and angular bandwidth $\Delta_{\theta}$, and a scene with transmission matrix $\tc$, Gkioulekas et al.~\cite{gkioulekas2015micron} prove that,
\begin{align}
    \bc\paren{x, l} = &\exp\paren{i\bar{\wn}\paren{d\paren{x}-l}} \temporal_{\Delta_{\wn}}\paren{d\paren{x}-l} \nonumber \\
	&\int_{x^\prime} \tc\paren{x, x^\prime} \spatial_{\Delta_{\theta}}\paren{x-x^\prime}\ud x^\prime,
    \label{eq:correlation}
\end{align}
where $\temporal_{\Delta_{\wn}}$ and $\spatial_{\Delta_{\theta}}$ are the \emph{temporal} and \emph{spatial coherence functions}, respectively,
\begin{equation}
	\temporal_{\Delta_{\wn}}\!\paren{\tau}\!\equiv\!\exp\!\paren{-\nicefrac{\paren{\Delta_{\wn}\tau}^2}{2}},\, \spatial_{\Delta_{\theta}}\!\paren{\epsilon} \!\equiv\! \sinc\!\paren{2\epsilon\bar{\wn}\Delta_{\theta}}.\label{eq:cf}
\end{equation}
We observe that the temporal and spatial coherence functions are shift-invariant kernels, with widths equal to the \emph{temporal} and \emph{spatial} coherence lengths $L_\temporal$ and $L_\spatial$,
\begin{equation}
	L_\temporal \equiv \nicefrac{1}{\Delta_{\wn}},\quad L_\spatial \equiv \nicefrac{1}{\bar{\wn}\Delta_{\theta}}.\label{eq:cl}
\end{equation}
The temporal and spatial coherence lengths are inversely proportional to the illumination's spectral and angular bandwidths $\Delta_{\wn}$ and $\Delta_{\theta}$, respectively. As the source becomes more monochromatic and collimated, the temporal and spatial coherence lengths become longer, and the temporal and spatial coherence functions become wider, respectively. 

Let us consider the case when the incident illumination is sufficiently temporally and spatially incoherent. Then, the spectral and angular bandwidths $\Delta_{\wn}$ and $\Delta_{\theta}$ are large enough that we can approximate:
\begin{equation}
	\temporal_{\Delta_{\wn}}\paren{\tau} \approx \delta\paren{\tau},\quad\spatial_{\Delta_{\theta}}\paren{\epsilon}\approx \delta\paren{\epsilon}.
\end{equation}
Consequently, the correlation in Equation~\eqref{eq:correlation} becomes
\begin{equation}\label{eq:impulse}
	\bc\paren{x, l} \approx \tc\paren{x, x}\delta\paren{d\paren{x}-l}.
\end{equation}
We observe that we measure non-zero correlation only when the position $l$ of the reference mirror matches the scene depth $d\paren{x}$. Then, the correlation equals the wave due to the direct-illumination-only response of the scene. In practice, there will be significant non-zero correlation only when the depth difference between reference and scene is smaller than the temporal coherence length, $\abs{l-d\paren{x}}< L_\temporal$. Likewise, the correlation will include contributions from elements of the transmission matrix close to the diagonal, corresponding to indirect light paths starting at scene points $x^\prime$ whose distance from the imaged point $x$ is smaller than the spatial coherence length, $\norm{x^\prime-x} < L_\spatial$. As we show in Section~\ref{sec:sunlight}, using sunlight corresponds to temporal and spatial coherence lengths in the order of $\unit[10]{\upmu m}$, suggesting that the approximation of Equation~\eqref{eq:impulse} is accurate.

    \vspace*{-5pt}
\section{Direct-only time-of-flight imaging}\label{sec:processing}
\vspace*{-5pt}

We now use the results of Section~\ref{sec:background} to derive a procedure for performing direct-only time-of-flight imaging using sunlight interferometry. We assume that we want to scan a depth range equal to $D$. We discretize this depth range to a set of axial locations $l_{m+1} = l_m + \nicefrac{L_\temporal}{2},\, m=1, \dots, M$, where $M=\ceil{\nicefrac{2D}{L_{\temporal}}}$. The axial separation between consecutive axial locations is due to the fact that, following the discussion at the end of the previous section, the temporal coherence length $L_\temporal$ determines the resolution at which we can measure differences between reference and scene depth.

We use the translation stage of our setup to move the reference mirror to each of the axial locations $l_m$. At each location, we use the camera to capture an image $I\paren{x,l_m}$---as we use a two-dimensional sensor, we acquire each such image for all locations $x$ in the field of view in a single exposure. We assume that, from each image $I\paren{x,l_m}$, we can isolate the corresponding correlation $\bc\paren{x,l_m}$ term in Equation~\eqref{eq:interf-def}. We detail how to do this later in this section. 

Then, from Equations~\eqref{eq:correlation} and~\eqref{eq:impulse}, the function
\begin{equation}\label{eq:tau}
    \tau\paren{x,l_m} \equiv \abs{\bc\paren{x,l_m}}^2,
\end{equation}
is approximately equal to the \emph{direct-only transient response} of the scene~\cite{o2014temporal,gkioulekas2015micron}---up to a change of variables $l = c\cdot t$ converting optical pathlength $l$ to time $t$ through the speed of light $c$. The approximation is due to the non-zero extent of the temporal and spatial coherence functions in Equation~\eqref{eq:correlation}, proportional to the temporal and spatial coherence lengths. We quantify these coherence lengths, and thus the resolution limits of our technique, in Section~\ref{sec:sunlight}.

\begin{figure}
    \centering
    \includegraphics[width=0.95\linewidth]{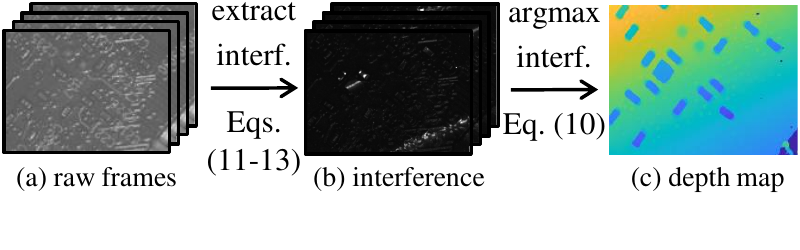}
    \vspace*{-25pt}
    \caption{\textbf{Processing pipeline.} \textbf{(a)} We capture a stack of images with the reference arm placed at a dense set of positions. \textbf{(b)} We blur the stack temporally and spatially, to estimate interference and correlation amplitude. \textbf{(c)} We extract a depth map by detecting the axial location maximizing correlation amplitude.}
    \label{fig:pipeline}
\end{figure}

We then estimate depth for each point in the scene as:
\begin{equation}\label{eq:depth}
    d\paren{x} = \argmax\nolimits_{l_m} \tau\paren{x, l_m}.
\end{equation}
We also estimate the direct-only intensity as $\tau\paren{x, d\paren{x}}$.

We note that Equation~\eqref{eq:depth} assumes that the direct-only transient has a single peak, corresponding to a single reflecting surface at each location $x$. This assumption is violated, e.g., when the camera observes a semi-transparent object in front of another object. However, in practice, our procedure can handle such cases without issue: As we record the entire transient function $\tau\paren{x, l_m}$, we can search for multiple peaks and detect the depth of all reflecting surfaces. We show this experimentally in Section~\ref{sec:results}.%
\footnote{We implicitly made the same assumption in Equation~\eqref{eq:correlation}, when we assumed that the terms depending on $l$ can be moved outside the integral. This corresponds to assuming that the scene's temporally-resolved transmission matrix has the form $\tc\paren{x,x^\prime,l}=\tc\paren{x,x}\delta\paren{l-d\paren{x}}$. We made this assumption to simplify exposition, but it is not necessary for either our theory or our imaging procedure to work.}

\begin{figure*}[t]
    \centering
    \includegraphics[width=0.95\textwidth]{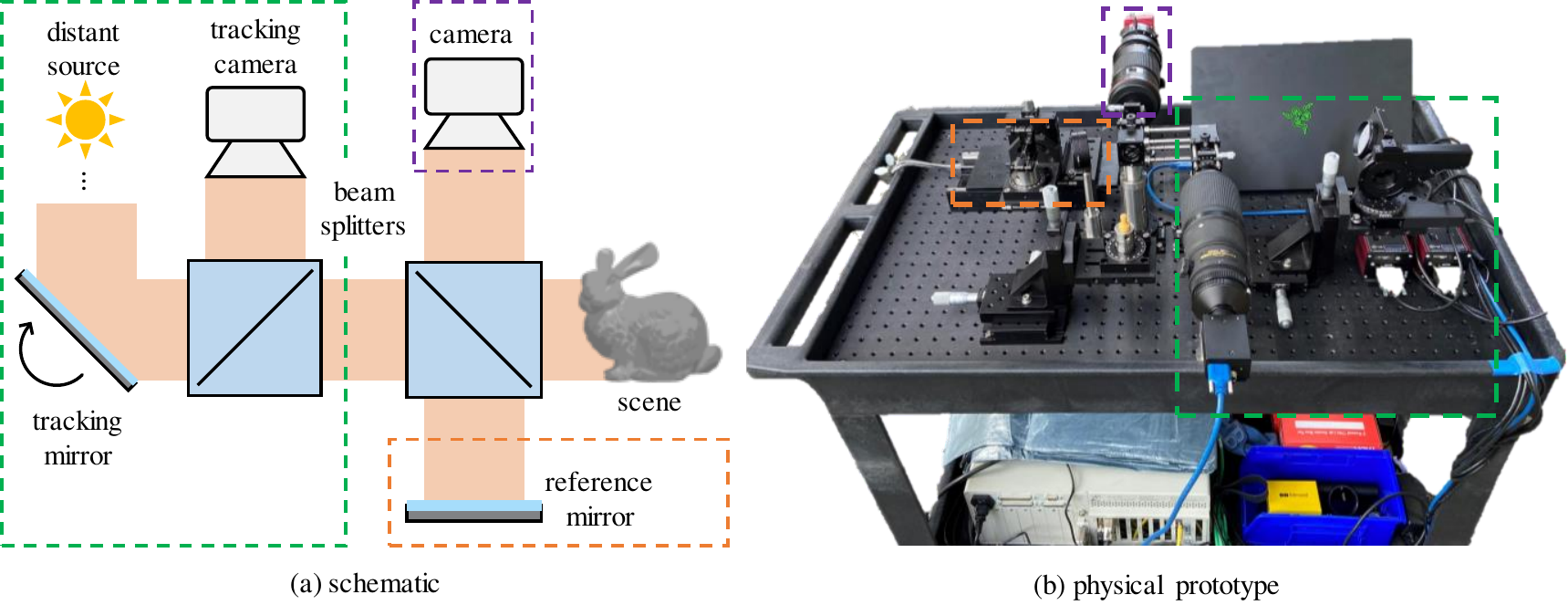}
    \vspace*{-10pt}
    \caption{\textbf{(a)} Schematic and \textbf{(b)} physical prototype of the sunlight interferometry setup on a utility cart.
    }\vspace*{-15pt}
    \label{fig:setup}
\end{figure*}

\boldstart{Relationship to optical coherence tomography.} The process we described above is reminiscent to \emph{time-domain optical coherence tomography (TD-OCT)} ~\cite{fercher2003optical,huang1991optical}. TD-OCT systems use a Michelson interferometer setup, and create incident spatially and temporally-incoherent illumination by placing a source such as an LED or halogen lamp at the focal plane of an illumination lens~\cite{xiao2016full,gkioulekas2015micron}.
Compared to TD-OCT, which requires active illumination, we show that we can perform time-of-flight imaging passively, using sunlight as temporally and spatially incoherent illumination.


\boldstart{Computing correlation.} We now describe how to compute the squared correlation amplitude $\tau\paren{x,l_m}$ from the measured axial stack of images $I\paren{x, l_m}, m=1,\dots,M$. The presence of the complex exponential in Equation~\eqref{eq:correlation} suggests that we can use phase shifting~\cite{deGroot2011} to estimate $\tau\paren{x,l_m}$. However, this would require shifting the reference mirror by fractions of the optical wavelength $\bar{\lambda}=\nicefrac{1}{\bar{\wn}}$ and capturing 3-4 images around \emph{each} location $l_m$. This makes acquisition prohibitively long, and is also impractical when operating outdoors---environment vibrations make it difficult to accurately perform sub-wavelength shifts.

Instead, we use the procedure proposed by Gkioulekas et al.~\cite{gkioulekas2015micron} to approximate the squared correlation amplitude $\tau\paren{x,l_m}$: First, we estimate the interference-free image $\bi\paren{x, l_m}$ by averaging nearby frames in the axial scan,
\begin{equation}
    \bi\paren{x,l_m} \approx \nicefrac{1}{N} \sum\nolimits_{n=m-N/2}^{N/2} I\paren{x, l_n}.
\end{equation}
Then, we estimate the squared interference as
\begin{equation}
    \!\!\!\!\!\real\!\big\{\bc\paren{x,l_m}\!\big\}^2\!\!\approx\! \br\!\paren{x, l_m}\!=\!\nicefrac{1}{4}\!\paren{I\!\paren{x, l_m}\!-\!\bi\!\paren{x, l_m}}^2.
\end{equation}
Lastly, we estimate the squared correlation amplitude as
\begin{equation}\label{eq:blurring}
    \tau\paren{x, l_m} = \paren{\mathcal{G}_s \ast \br}\paren{x, l_m},
\end{equation}
where $\mathcal{G}_s$ is a Gaussian kernel with standard deviation of $s$ pixels. We visualize our pipeline for correlation estimation in Figure~\ref{fig:pipeline}, and provide pseudocode in the supplement.

The intuition behind the blurring operation in Equation~\eqref{eq:blurring} is as follows: When imaging scenes with rough surfaces, the interference appears as speckle. This speckle is created due to pseudo-random sub-wavelength pathlength shifts, accounting for the surface's microstructure. Blurring (squared) interference speckle around some location is approximately equal to averaging squared interference measurements at sub-wavelength shifts. This is how phase shifting techniques estimate squared correlation amplitude~\cite{deGroot2011}. 
    \begin{figure}[t]
    \centering
    \includegraphics[width=0.95\linewidth]{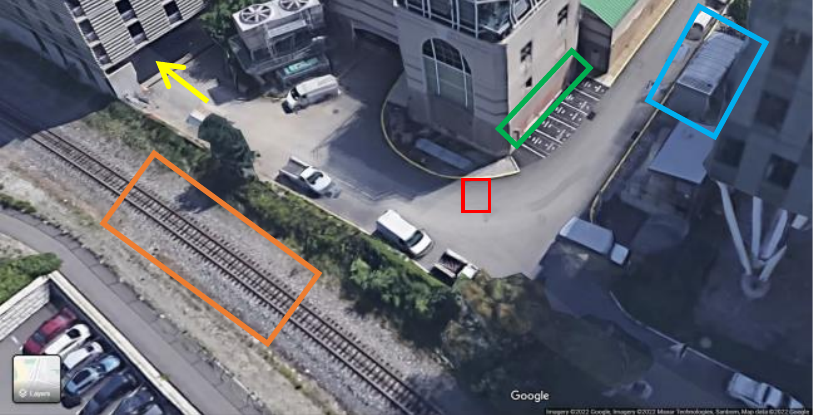}
    \vspace*{-5pt}
    \caption{\textbf{Google Maps view of experimental site.} This image showcases the challenging conditions of our experiments. \textbf{Red}: setup location. \textbf{Yellow}: parking garage, causing constant traffic with multiple vehicles (including trucks) passing per acquisition. \textbf{Orange}: train tracks, with trains at frequent intervals. \textbf{Green}: building air intake causing a constant loud background hum. \textbf{Blue}: power transformer also causing a loud constant background hum.}
    \label{fig:location}
    \vspace*{-10pt}
\end{figure}

\begin{figure}[t]
    \centering
    \includegraphics[width=0.95\linewidth]{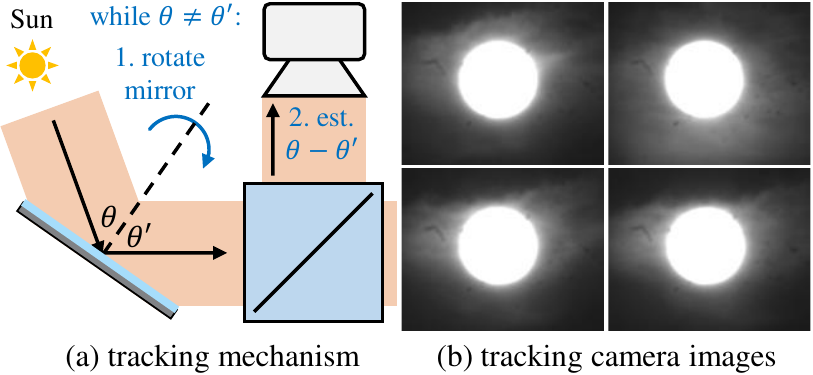}
    \vspace*{-10pt}
    \caption{\textbf{Sun tracking.} \textbf{(a)} In a closed loop, we estimate the direction of the light beam reflecting from the mirror with a tracking camera focused at infinity, and correct it by rotating the mirror. \textbf{(b)} Tracking camera images of the Sun during a centering operation.}
    \label{fig:sun-tracking}
\end{figure}
\vspace*{-15pt}
\section{Hardware prototype}\label{sec:setup}
\vspace*{-5pt}
We build an experimental prototype to demonstrate the capabilities of sunlight interferometry. We show an image of our prototype in Figure~\ref{fig:setup}. We discuss here some key implementation aspects, and refer to the supplement for full implementation details, including a parts list, construction, alignment, and calibration procedures.

\boldstart{Key specifications.} Our prototype uses a camera equipped with a CCD sensor (pixel pitch $\unit[3.7]{\upmu m}$, $3400\times 2700$ pixels) and a compound lens (focal length $\unit[300]{mm}$). We use a reproduction ratio of 1:1, resulting in a field of view of $\unit[12.5]{mm}\times\unit[10]{mm}$, and working distance of $\unit[600]{mm}$. For axial scanning, our prototype uses a translation stage with resolution $\unit[10]{nm}$. We use an axial stepsize of $\unit[5]{\upmu m}$ (Section~\ref{sec:sunlight}) and capture 1000 images per scan at exposure times of $\unit[50]{ms}$. This results in acquisition times around a minute per scene. We follow Gkioulekas et al.~\cite{gkioulekas2015micron} in setting the various imaging settings to maximize interference contrast, and thus signal-to-noise ratio (SNR).

\boldstart{Tracking the Sun.} As it is not practical to orient the optical axis of the interferometer towards the Sun, and to account for motion of the Sun during acquisition, we equip our setup with a custom Sun-tracking assembly (Figure~\ref{fig:setup}, green boxes). This assembly comprises a mirror whose 3D orientation is controlled via two motorized rotation stages. We set the mirror orientation using information from a tracking camera separate from the imaging camera: We set the camera so that its optical axis is parallel to that of the interferometer, and its lens focused at infinity. The camera directly images the Sun through the mirror, and provides feedback (through a digital PID controller) to the rotation stages to adjust the mirror orientation, so that the Sun remains at the center of the camera's field of view.

\boldstart{Outdoor operation.} To perform experiments outdoors and mitigate the vibration effects, we build our setup on an optical breadboard mounted on a utility cart. The imaged scenes are on the ground or a tripod. We do not use any enclosures, and both our setup and  scenes are fully exposed to ambient light. Figure~\ref{fig:location} shows our experimental site. It includes strong sources of environment noise, including moving vehicles, air, and ambient sound. These experimental conditions are in stark contrast with those in prior works on interferometric computational imaging~\cite{gkioulekas2015micron,Kotwal2020,li2018sh,li2017high,cossairt2014,maeda2018dynamic}, which require carefully-controlled environments (e.g., dark room, vibration-isolated optical tables, no air flow).

\boldstart{Cloudy sky conditions.} We performed most of our experiments under direct sunlight. However, we found that under scattered or very high-altitude clouds, we are still able to obtain high-quality depth, even if the clouds are fast-moving. Figure~\ref{fig:sun-tracking}(b) shows frames acquired by the tracking camera during a single acquisition, showing the tracked Sun under fast-moving, scattered cloudy conditions.

\boldstart{Cost discussion.} Our design trades off the light source of an active system for the Sun tracking module of a passive system. We expect that, in a carefully designed implementation, the passive system will have a significantly lower cost than the active one. The parts we used for tracking in our prototype were simply parts we had readily access to, and can be replaced with much cheaper off-the-shelf components: For example, we can replace the rotation stages with an a pair of galvo mirrors, and the machine vision tracking camera with an inexpensive camera (e.g., Raspberry Pi HQ Camera). Considering the high cost of a light source suitable for OCT (including drivers for current control and temperature stabilization), replacing the source with such a tracking unit reduces overall cost.

\boldstart{Comparison to using an active source.} In addition to improved mobility, form factor, power consumption (Section~\ref{sec:intro}) and cost (above), using passive sunlight instead of an active light source for outdoor operation provides more flexible control of imaging performance characteristics. In particular, we can increase axial resolution simply by increasing the bandwidth of the spectral filter we use, without impacting power consumption or signal-to-background ratio---the ratio of sunlight steered through our system to ambient sunlight is unaffected by the filter. By contrast, when using an active light source, it is necessary to also increase the source emission bandwidth to match the increased filter bandwidth. 
If the source's power remains the same, the result is worse signal-to-background ratio---more ambient light is let through the filter, while active light remains the same. Countering this requires increasing active light, which results in increased power consumption.

\begin{figure*}[t]
    \centering
    \includegraphics[width=\textwidth]{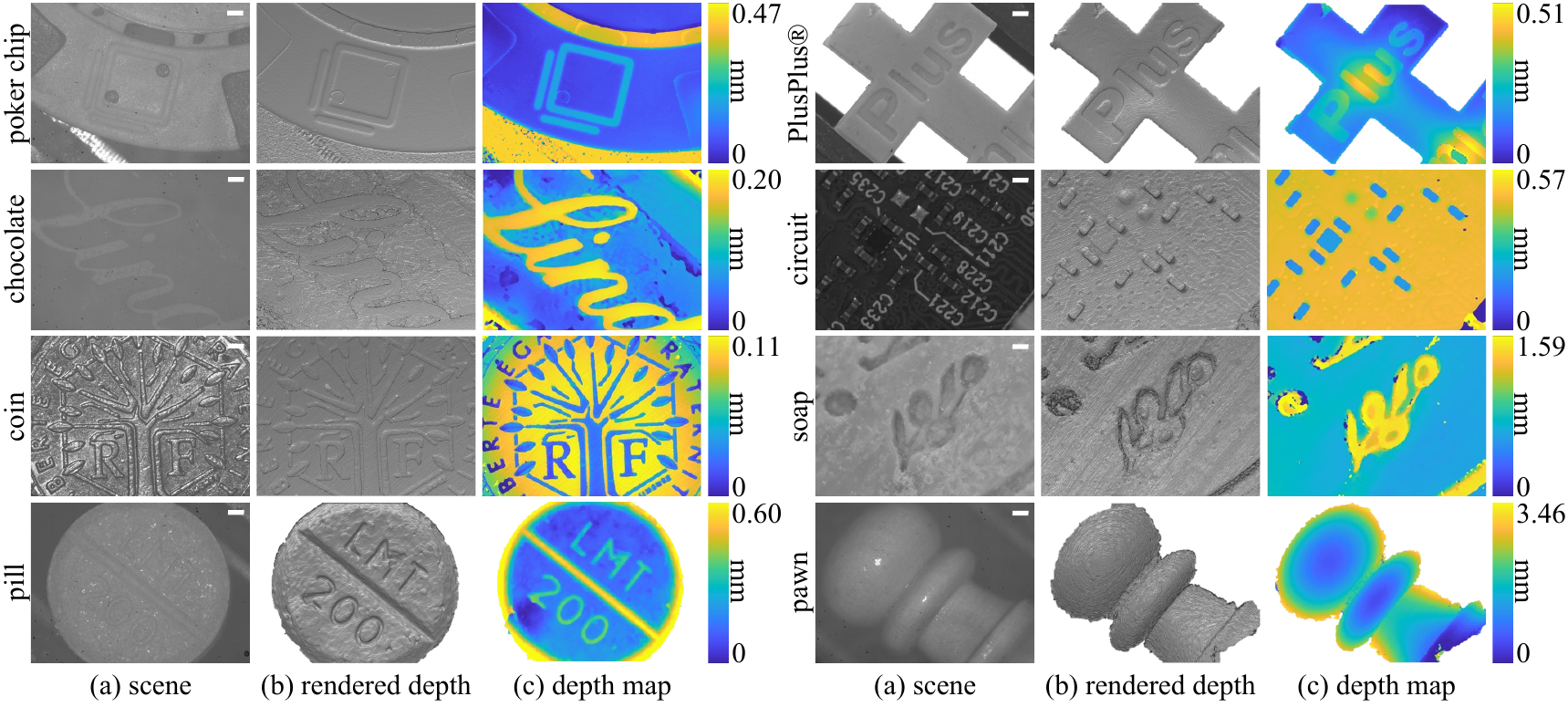}
    \vspace*{-20pt}
    \caption{\textbf{Depth reconstructions for a variety of challenging scenes using sunlight.} The challenges include environmental factors (Figure~\ref{fig:location}) and low signal quality (detailed in the text). The white bars on the scene images indicate one millimeter of lateral size.}
    \label{fig:results-sunlight}
    \vspace*{-15pt}
\end{figure*}
    \vspace*{-5pt}
\section{Coherence properties of filtered sunlight}\label{sec:sunlight}
\vspace*{-5pt}

We measure the temporal and spatial coherence lengths of sunlight, to determine the resolution limits of sunlight interferometry, and inform our experiments in Section~\ref{sec:results}.


%
\begin{figure}
    \centering
    \includegraphics[width=0.95\linewidth]{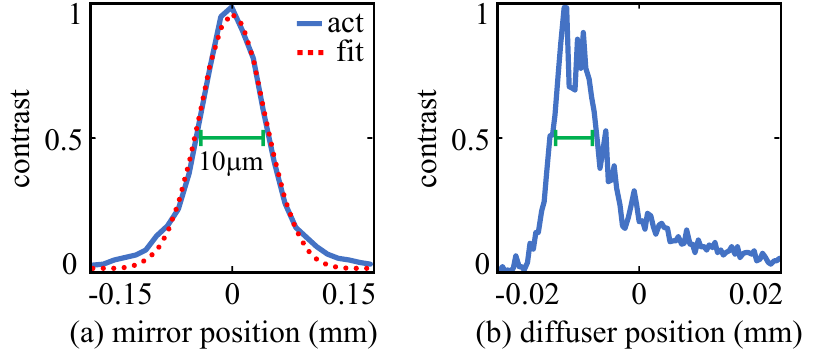}
    \vspace*{-10pt}
    \caption{\textbf{Temporal coherence of spectrally-filtered sunlight.}
    We perform Gaussian fits to measurements for both mirror and planar diffuser scenes. From their full-width half-maximum, we estimate temporal coherence length $L_\temporal \approx \unit[10]{\upmu m}$.}
    \vspace*{-10pt}
    \label{fig:temporal-coherence}
\end{figure}

\boldstart{Temporal coherence length.} Previous work has estimated the temporal coherence length of direct sunlight to be approximately $L_{\temporal} = \unit[0.3]{nm}$~\cite{ricketti2022coherence}. However, performing sunlight interferometry at these axial resolutions using the full spectrum of sunlight is impractical: First, chromatic aberrations from optics (lenses, beamsplitters) when performing interferometry with very broadband illumination can drastically reduce interference contrast~\cite{fercher2003optical}. Second, environment vibrations when operating an interferometer outdoors are in the order of hundreds of micrometers. Together, these two factors result in impractically low SNR.

Instead, we limit spectral bandwidth using a spectral filter with central wavelength $\unit[550]{nm}$ and bandwidth $\unit[20]{nm}$. To measure the resulting temporal coherence length, we perform two experiments where we place either a mirror or a planar diffuser at the scene arm of our setup. For each scene, we measure $\tau$ at a dense set of reference axial locations. From Equations~\eqref{eq:correlation}-\eqref{eq:cl}, we expect the square root of the measurements to be shaped like a Gaussian with standard deviation equal to the temporal coherence length. We perform a Gaussian fit to the measurements, and estimate the temporal coherence length as the fit's full-width half-maximum. Figure~\ref{fig:temporal-coherence} shows the results. For both mirror and planar diffuser, we estimate the temporal coherence length, and thus axial resolution, to be approximately $L_{\temporal} = \unit[10]{\upmu m}$.

\begin{figure}
    \centering
    \includegraphics[width=0.95\linewidth]{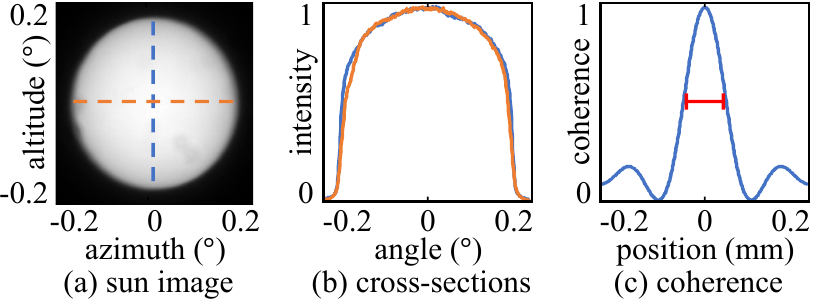}
    \vspace*{-10pt}
    \caption{\textbf{Spatial coherence length of sunlight.}
    \textbf{(a-b)} An image of the Sun by the tracking camera yields an angular size of $0.57^\circ$. \textbf{(c)} From this, we estimate spatial coherence length $L_\spatial\approx\unit[100]{\upmu m}$.}
    \label{fig:spatial-coherence}
\end{figure}


\boldstart{Spatial coherence length.} To measure the spatial coherence length of sunlight, we use the tracking camera of our setup to capture a direct image of the Sun. At infinity focus, the spatial extent of the Sun in this image is directly proportional to the angular extent of sunlight (up to the focal length of the lens). We can then use Equations~\eqref{eq:cf}-\eqref{eq:cl} to estimate the spatial coherence length. Figure~\ref{fig:spatial-coherence} shows the results. We estimate the angular extent of sunlight to be around $0.57^\circ$, corresponding to a spatial coherence length of approximately $L_{\spatial}=\unit[100]{\upmu m}$. This is consistent with theoretical and experimental estimates in prior work~\cite{Agarwal04,Mashaal12}.


    \vspace*{-5pt}
\section{Experiments}\label{sec:results}
\vspace*{-5pt}

\boldstart{Depth scanning}. In Figure~\ref{fig:results-sunlight}, we show scans for variety of scenes. These feature micrometer-scale geometric details, and pose a variety of challenges including low reflectivity (circuit board), specularities (coin, pawn) and strong subsurface scattering (chocolate, soap, pill). We obtain high quality reconstructions, despite operating under very adverse environment conditions (Section~\ref{sec:setup}).

\begin{figure}[t!]
    \centering
    \includegraphics[width=0.95\linewidth]{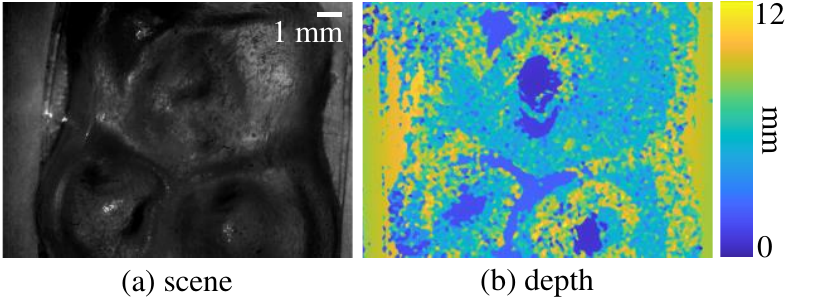}
    \vspace*{-10pt}
    \caption{\textbf{Depth reconstruction under extremely low SNR.} The gummy bear has very little backreflected direct light, making it extremely challenging to capture depth.}
    \label{fig:results-challenging}
\end{figure}

%
\begin{figure}[t!]
    \centering
    \includegraphics[width=\linewidth]{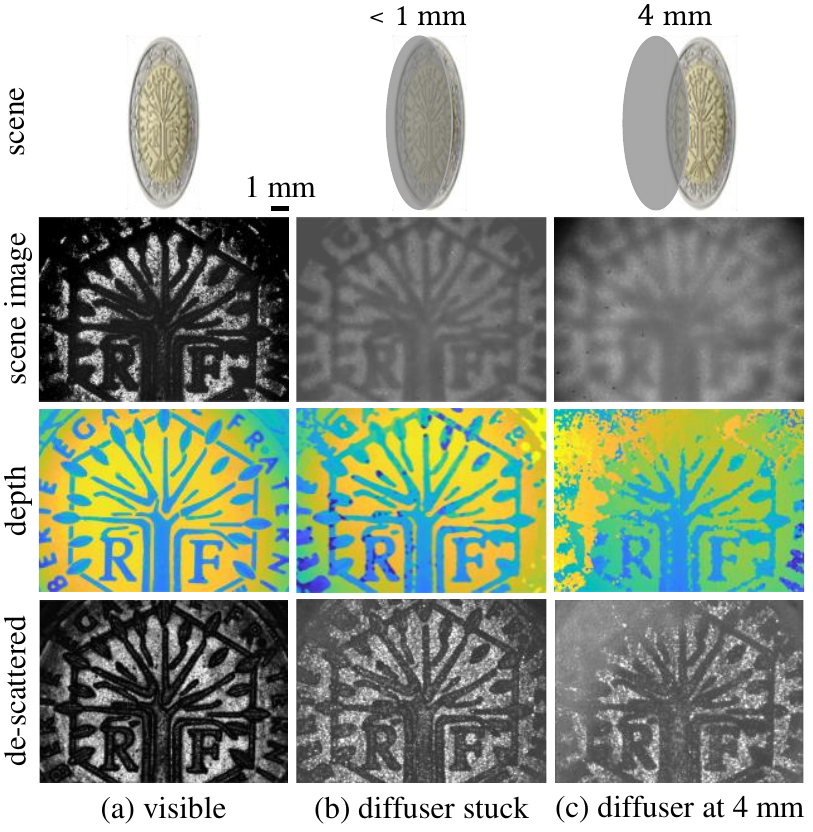}
    \vspace*{-20pt}
    \caption{\textbf{Seeing through diffusers.} We image a metallic coin under three conditions: \textbf{(a)} directly, \textbf{(b)} with a ground glass diffuser stuck to it and \textbf{(c)} with the diffuser placed $\unit[4]{mm}$ away. The images of the scene in \textbf{(b-c)} are blurred due to scattering in the diffuser. The interference component rejects scattering.}
    \label{fig:descattering}
    \vspace*{-10pt}
\end{figure}

\boldstart{Scanning under extremely low SNR.} A key challenge in sunlight interferometry is the very low SNR in captured images. This is for two reasons: First, in Equation~\eqref{eq:interf-def}, the interference is typically much smaller than the interference-free image. Second, the scenes are illuminated not only by the sunlight directed at them by the interferometer, but also by the ambient sunlight, further reducing interference contrast. To test the SNR limits of our technique, in Figure~\ref{fig:results-challenging}, we scan a semi-transparent object (gummy bear) that backreflects very little light. Despite this, our technique still acquires accurate depth for most of the scene, except for parts observed at near-grazing angles (no backreflected light).

\boldstart{Seeing through diffusers.} To test the ability of our technique to isolate direct-only (ballistic) light and measure multiple depth peaks, in Figure~\ref{fig:descattering}, we scan a scene where a coin is occluded from the camera by a ground glass diffuser. The conventional images of the scene show strong blur because of the diffuser. Our technique can accurately acquire the depth \emph{and} clean images of the occluded coin. 


%
\begin{figure}[t!]
    \centering
    \includegraphics[width=\linewidth]{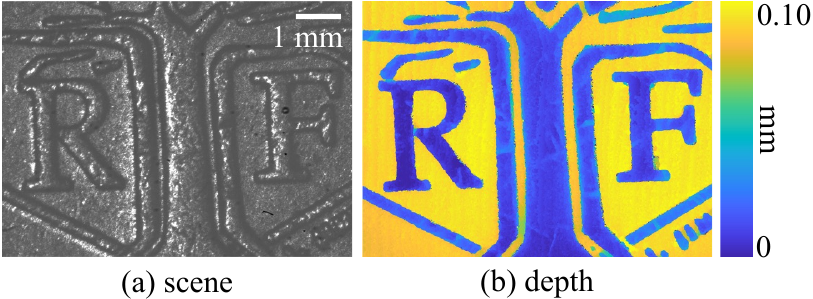}
    \vspace*{-20pt}
    \caption{\textbf{Depth reconstruction with indoor illumination} at a distance of about five feet from the interferometer.}
    \label{fig:results-indoor}
\end{figure}

\boldstart{Time-of-flight with indoor illumination.} Even though we focus on sunlight interferometry, Figure~\ref{fig:results-indoor} shows an experiment testing the ability to perform time-of-flight imaging with passive \emph{indoor} illumination. For this, we point the Sun tracking module to a ceiling light about five feet from the tracking mirror. We scan the same metallic coin as in Figure~\ref{fig:results-sunlight}, which backreflects a lot of light, resulting in accurate depth recovery. However, this scanning setting remains challenging for non-metallic objects, because of the very low SNR between backreflected versus ambient light.

    \vspace*{-5pt}
\section{Limitations and conclusion} 
\vspace*{-5pt}

\boldstart{Mechanical scanning and comparison to Fourier-domain and swept-source OCT.} As we mentioned before, the operation of our technique is similar to that of (active illumination) time-domain OCT techniques. Both techniques require mechanical scanning along the axial dimension. This results in relatively slow acquisition times (around a minute for our scenes), and increased mechanical vibrations that reduce resolution and SNR. In OCT, these issues are ameliorated through the use of alternative technologies that eliminate the need for axial scanning, such as swept-source and Fourier-domain OCT. Swept-source OCT requires sweeping the wavelength of a monochromatic source, which is not possible when working with sunlight. Fourier-domain OCT uses a spectrometer to measure interference as a function of wavelength, which relates to the transient response through a Fourier transform. Unfortunately, as two-dimensional spectrometers of sufficiently high spectral resolution are not available, this requires using raster scanning to acquire full-frame images. Additionally, most implementations of Fourier-domain OCT used single-mode fiber optics, which are incompatible with spatially-incoherent sunlight. The development of Fourier-domain sunlight interferometry is an exciting future direction.

\boldstart{General light transport probing.} Our method relies on the spatial incoherence of sunlight to capture direct-only measurements of the scene. However, modern computational light transport imaging systems are capable of much more general forms of light transport probing, which currently remain out of reach for sunlight interferometry. Kotwal et al.~\cite{Kotwal2020} showed that, by modulating the spatial and temporal coherence properties of the incident illumination, it is possible to implement very general forms of probing inside interferometric imaging systems. Similar forms of modulation can, conceivably, be incorporated in sunlight interferometry. In turn, this would make it possible to realize \emph{passive} variants of computational light transport techniques such as focusing inside scattering and non-line-of-sight imaging. We hope that our work will motivate and facilitate future research towards these fascinating possibilities.

\boldstart{Conclusion.} We have demonstrated a passive interferometric method for micron-scale depth sensing outside the controlled environment of an optics lab. Using light from ambient sources such as the Sun removed the need to use expensive, power-hungry, high-power or eye-unsafe lasers for interferometry. The inherent incoherence properties of sunlight made it possible to acquire time-of-flight information at micrometer scales, and insensitive to indirect illumination effects that confound depth inference. Additionally, engineering a robust optical setup mitigated most of the detrimental effects associated with performing experiments in uncontrolled outdoor environments, such as vibrations and wind, that typically hinder interferometric measurements. Together, these contributions allowed us to demonstrate, for the first time, passive imaging capabilities such as micrometer-scanning of challenging scenes, direct-only imaging, and imaging through diffusers. We also take first steps towards enabling these capabilities under passive \emph{indoor} illumination. We hope that our work will motivate applications in areas such as inspection, fabrication, robotic grasping, and biomedical imaging.

\boldstart{Acknowledgments.} We thank Sudershan Boovaraghavan and Yuvraj Agrawal, who provided the samples for some of our experiments. This work was supported by NSF awards 1730147, 2047341, 2008123 (NSF-BSF 2019758), and a Sloan Research Fellowship for Ioannis Gkioulekas.

    \appendix
    \section{Implementation details} \label{app:implementation}
We discuss implementation details of our sunlight interferometry setup (replicated in Figure~\ref{fig:setup}), and provide a full parts list in Table~\ref{table:parts}. We take inspiration from the interferometer designs by Gkioulekas et al.~\cite{gkioulekas2015micron} and Kotwal et al.~\cite{Kotwal2020}, and modify them to use sunlight and achieve robustness to environmental conditions for outdoor operation. For alignment of optical components, we use the methods described by Gkioulekas et al.~\cite{gkioulekas2015micron} and Kotwal et al.~\cite{Kotwal2020}.

\boldstart{Vibration isolation.} The factor most detrimental to the quality of our results is vibrations in uncontrolled outdoor conditions. Vibrations arise from multiple sources: (a) mechanical vibrations from the ground that propagate up the cart because it has no suspension; (b) movement of the optical components on the table with respect to each other; and (c) lateral and axial movement of the scene and reference and tracking mirrors induced by strong winds.

To mitigate (a), we place our setup on a passively damped honeycomb optical breadboard from Thorlabs on the upper level of our utility cart. We found in experiments that using the breadboard and eliminating any other sources of vibrations (such as the computer in Figure~\ref{fig:setup}(b) and cables connected to the various components) on the upper level of the cart is sufficient. We tightly route the cables so they do not move parts of the setup. With the damped breadboard, vibrations from the lower level of the cart do not travel to the optical setup, so any noisy equipment such as the reference arm translation stage controller (the white cabinet in Figure~\ref{fig:setup}(b)) can be placed on the lower level.

To mitigate (b), we modified the interferometer design from Kotwal et al.~\cite{Kotwal2020} so as to tightly clamp down the optical components to each other and to the breadboard. We achieved this by connecting the beamsplitters and the cameras with a $\unit[30]{mm}$ cage system using components from Thorlabs. It is not practical to connect the tracking and reference mirrors similarly, so we clamped them tightly to the breadboard using mounting components from Thorlabs.


\begin{figure*}
    \centering
    \includegraphics[width=\textwidth]{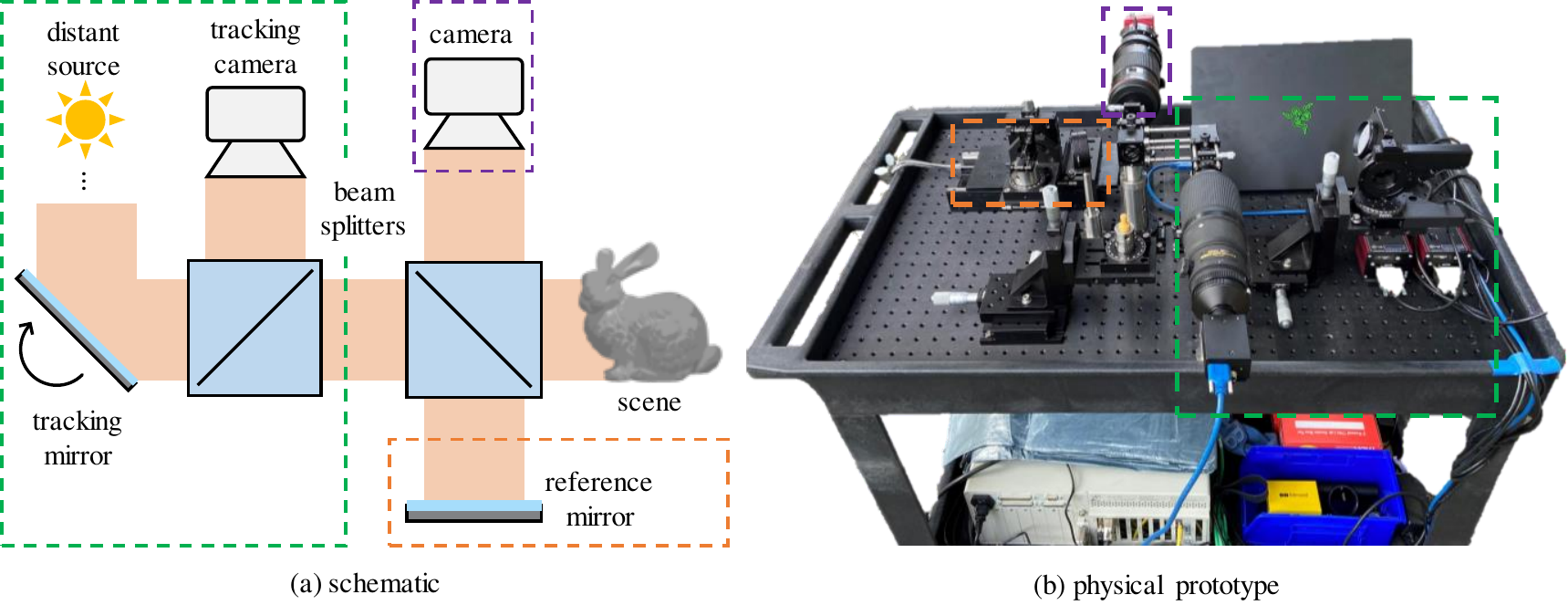}
    \vspace{-20pt}
    \caption{\textbf{(a)} Schematic and \textbf{(b)} physical prototype of the sunlight interferometry setup on a utility cart.
    }
    \vspace*{-15pt}
    \label{fig:setup}
\end{figure*}

We can mitigate mirror movement in (c) by choosing high-resistance translation stages and high-torque rotation stages. We discuss the properties of our stages below. We found that 
our setup can tolerate winds as fast as \unit[25]{mph}. 

We estimated the amplitude of vibrations on the setup surface by observing the movement of fringe patterns when we mount mirrors on both scene and reference arms. When we built our setup on an undamped optical breadboard on the cart, we measured vibration amplitude to be around \unit[10]{\textmu m}. By contrast, when we used a passively-damped breadboard, vibration amplitude was below a micrometer (though still orders of magnitude larger than what we measured indoors on an optical table with pneumatic isolation). For the coherence lengths and exposure times we use, we found empirically that this was sufficiently small to allow high-quality depth sensing.


\boldstart{Tracking the Sun.} As we mention in the main paper, it is important to track the Sun and center the sunlight beam along the optical axis of the system several times during acquisition. To keep the distribution of sunlight across the field of view uniform, we found that it is necessary to re-center the beam every 100 intensity images. In the duration of capturing 100 images at $\unit[50]{ms}$ per image, the Sun moves \unit[0.02]{$^\circ$} in the sky. We perform the tracking using two motorized precision rotation stages from Thorlabs, one for the azimuth and one for the altitude of the mirror. These stages have minimum incremental motions of $\unit[0.03]{^\circ}$, making them perfect for our setup. We found in tests outdoors that the torque provided by these rotation stages was enough to keep the tracking mirror stable in winds as fast as $\unit[25]{mph}$. As the control system, we found that a proportional controller with $k_p$ set as half the sensitivity of pixel displacements to rotation stage angles was sufficient. 

\boldstart{Reference translation stage.} To ensure that we place the reference mirror at the desired positions accurately enough for micron-scale resolutions, we need a translation stage that has a minimum incremental translation less than one micron. We use the XMS160 translation stage from Newport that has a minimal incremental translation of $\unit[10]{nm}$. In addition, this stage guarantees low-noise and high-resistance operation, preventing loss of interference contrast due to reference mirror position noise caused by mechanical vibrations and flowing wind.



\boldstart{Beamsplitters.} We use thin plate beamsplitters from Thorlabs. The beamsplitter sending light to the tracking camera is a 10:90 (R:T) beamsplitter, reflecting 10\% of the input light to the tracking camera. The imaging beamsplitter is a 50:50 (R:T) beamsplitter. We choose plate beamsplitters over cube beamsplitters because cubes cause significant interreflections, and over pellicle beamsplitters because of their tendency to distort due to airflow.

\boldstart{Mirrors.} We use high-quality protected aluminum mirrors of guaranteed $\nicefrac{\lambda}{4}$ flatness to ensure a uniform phase distribution throughout the sunlight beam. 

\boldstart{Spectral filter.} As mentioned in the main paper, we control the spectral bandwidth of sunlight using a spectral filter. We choose the spectral filter to balance light efficiency, acquisition time, temporal coherence length, and signal to noise ratio. If we increase the spectral filter bandwidth, the acquisition time decreases, the signal to ratio increases, and the temporal coherence length decreases. The latter makes depth recovery more sensitive to vibrations from the environment. We found in experiments that a spectral filter with central wavelength $\unit[550]{nm}$ and bandwidth $\unit[20]{nm}$ works best with out setup.

\begin{table*}
    \centering
    \caption{List of major components used in the optical setup in Figure~\ref{fig:setup}(b).}
    \vspace*{-10pt}
    \resizebox{\linewidth}{!}{
    \begin{tabular}[center]{|l|c|c|c|}
        \hline {\bf description} & {\bf quantity} & {\bf model name} & {\bf company} \\
        \hline 3' $\times$ 2' utility cart & 1 & \url{https://www.amazon.com/dp/B001602VI2}& Rubbermaid \\
        \hline passively damped optical breadboard & 1 & B2436FX & Thorlabs \\
        \hline 2.56" motorized precision rotation stage & 1 & PRMTZ8 & Thorlabs \\
        \hline 1" motorized precision rotation stage & 1 & PRM1Z8 & Thorlabs \\
        \hline K-Cube brushed DC servo motor controller & 2 & KDC101 & Thorlabs \\
        \hline 2" round protected aluminum mirror & 1 & ME2-G01 & Thorlabs \\
        \hline 25 $\times$ 36 mm plate beamsplitter, 10:90 (R:T) & 1 & BSN10R & Thorlabs \\
        \hline 25 $\times$ 36 mm plate beamsplitter, 50:50 (R:T) & 1 & BSW10R & Thorlabs \\
        \hline 300 mm compound lens & 2 & AF Micro Nikkor 300mm 1:4 D IF-ED & Nikon \\
        \hline tracking camera & 1 & Grasshopper3 USB3 GS3-U3-41C6M-C & FLIR \\
        \hline imaging camera & 1 & Blackfly S USB3 BFS-U3-122S6M-C & FLIR \\
        \hline 1" round protected aluminum mirror & 1 & ME1-G01 & Thorlabs \\
        \hline 2" absorptive neutral density filter kit & 1 & NEK03 & Thorlabs \\
        \hline ultra-precision linear motor stage, 16 cm travel & 1 & XMS160 & Newport Corporation \\
        \hline ethernet driver for linear stage & 1 & XPS-Q2 & Newport Corporation \\
        \hline 550 $\pm$ 20 nm bandpass spectral filter & 1 & FB550-40 & Thorlabs \\
        \hline 
    \end{tabular}}
    \label{table:parts} 
\end{table*}

\boldstart{Alignment and calibration.} In addition to handling the factors detrimental to interferometric depth reconstruction that we detailed above, we need to deal with the imperfections within the system's construction. The stability that the cage system provides comes at the loss of the capability to finely adjust the orientations and positions of individual components. We need to calibrate for the errors caused by misalignment. In addition, the tracking and control system parameters are critical to set properly for Sun tracking to work fast enough without becoming unstable. With this in view, there are eleven important calibration parameters in the setup for accurate light delivery and tracking stability.
\begin{enumerate}
    \item Tracking camera principal point: Due to small misalignments in the cage system, when the tracking mirror deflects light perfectly parallel to the optical axis of the system, the image of the Sun at infinity does not appear at the center of the tracking damera sensor. We manually adjust the tracking mirrors to make sure that the shadows of the cage system rods on the beamsplitter mounts disappear. The position of center of the Sun image at this tracking mirror position is the principal point of the tracking camera.   
    \item Tracking motor speeds and accelerations: The tracking mirror rotation stages have controllable limits on angular speeds and accelerations. We tune these to achieve the minimum incremental motion at the principal point as fast as possible while still being stable.   
    \item Sensitivity of pixel displacements on the camera to tracking mirror angular displacements: To get a ballpark number for the proportional controller gain, we estimate this sensitivity by rotating the tracking mirror in both dimensions by the minimum incremental rotation of the tracking mirror stages and measure the displacement of the Sun image center.
    \item Proportional controller gains: Finally, the proportional controllers we use to output control signals to the rotation stages need to be as fast as possible without overshooting. Overshooting can be catastrophic, because a large overshoot leads the Sun out of the field of view of the tracking camera.
    \item Time between tracking adjustments: The movement of the Sun causes spatial intensity non-uniformities in the captured scan. To minimize this while considering the minimum incremental motion of the rotation stages, we calibrate how many scan images to capture between adjustments using the exposure time per image.
\end{enumerate}

\boldstart{Temperature conditions.} We conducted experiments over a wide range of temperatures: 3 \textdegree C -- 20 \textdegree C. We found that, at least in this temperature range, we did not need to recalibrate to account for temperature changes. Off-the-self optics and optomechanical components are generally designed for this temperature range, and therefore their properties remain stable as temperature changes.

    \section{Visualizations of direct light-in-flight}
\begin{figure*}
    \centering
    \includegraphics[width=\textwidth]{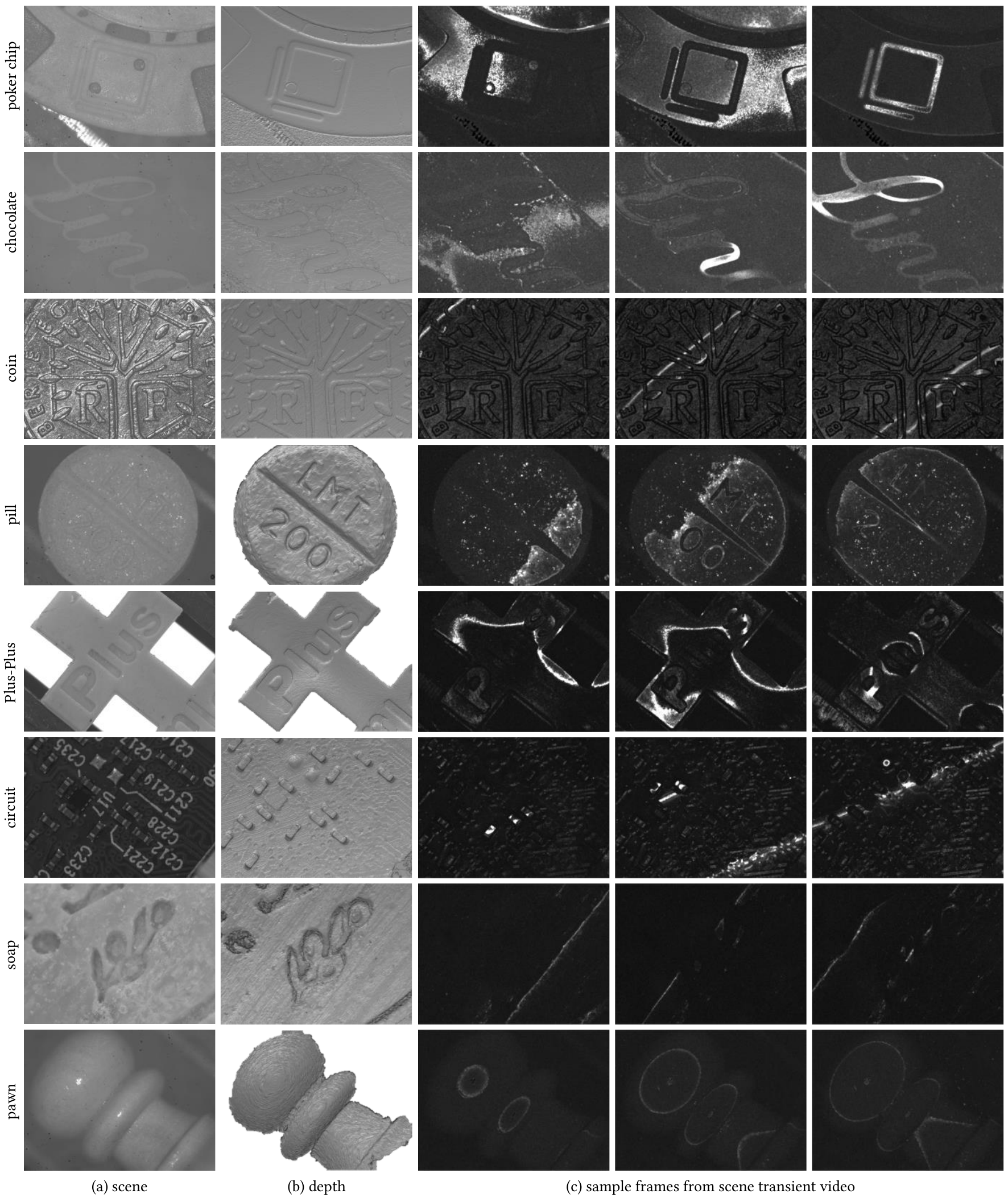}
    \caption{\textbf{(b)} Depth reconstructions using sunlight, rendered as 3D surfaces and \textbf{(c)} sample raw frames from the depth scans.}
    \label{fig:results-sunlight-renderings}
\end{figure*}

As Equation (9) from the main paper shows, the function $\tau\paren{x, l_m}$ we estimate from our depth recovery pipeline is the direct-only transient response of the scene to incoming illumination. We can then visualize slices of the propagation of direct-only light in the scene as a stack of images of $\tau$. We show some of these slices in Figure~\ref{fig:results-sunlight-renderings} for the scenes in Figure~7 of the main paper. 
    \section{Data and reconstruction code}
To facilitate reproducibility, we provide in the project website all of the data and scripts used to generate results in the main paper. In addition, Figure~\ref{fig:code-acquire} provides Matlab code for acquiring measurements with the optical setup of Figure~\ref{fig:setup}(b). Figure~\ref{fig:code-track} provides code for tracking the Sun during acquisition, and Figure~\ref{fig:code-process} provides code for recovering direct-only transients as explained in the post-processing section of the main paper.

\begin{figure*}[ht]
\begin{lstlisting}[language=Matlab,frame=single]
function frames = acquire(positions, imageSize, vid, refMotor, ...
                          trackingVid, trackingMotor, adjustEvery, ...
                          trackingSensitivity, trackingKp, trackingTol)
    
    % Acquire passive interferometry data

    % Reset motor  
    refMotor.goto(positions(1));
    resolution = positions(2)-positions(1);

    % Initialize array for frames
    frames = zeros([imageSize numel(positions)], 'uint16');

    % At every position
    for i = 1:numel(positions)
        
        % Center the sun at every adjustEvery positions
        if mod(i, adjustEvery) == 0
            track(trackingVid, trackingMotor, trackingSensitivity, trackingKp, trackingTol);
        end

        % Record an image at this position
        frames(:, :, i) = getsnapshot(vid);

        % Translate reference arm to the next position and give it time to settle
        refMotor.translate(resolution);
        pause(0.1);

    end

end
\end{lstlisting}
\caption{Matlab code for recovering depth from our measurements}
\label{fig:code-acquire}
\end{figure*}
\begin{figure*}[ht]
\begin{lstlisting}[language=Matlab,frame=single]
function track(vid, motor, sensitivity, kp, tol)

    % Center the sun on the frame

    % One-dimensional tracking for brevity
    % Easily extended to two dimensions
    err = Inf;
    correction = 0;

    % Loop unless you can tolerate error
    while abs(err) > tol

        % Translate tracking motor by previous correction
        motor.translate(correction);

        % Get sun center coordinates
        image = getsnapshot(vid);
        xx = getSourceCenter(image);

        % Correct based upon the error between the image center and sun center 
        err = xx-size(image)/2;
        correction = -kp*sensitivity*err;
    end

end
\end{lstlisting}
    \caption{Matlab code for sun tracking}
    \label{fig:code-track}
\end{figure*}
\begin{figure*}[ht]
\begin{lstlisting}[language=Matlab,frame=single]
function [depth, direct] = process(frames, positions, spatialWindow, temporalWindow)

    % Reconstruct depth from passive interferometry measurements
    % frames: HxWxN array of measurements from the camera
    % positions: set of positions of the reference arm
    % spatialWindow: blur kernel width for blurring speckle
    % temporalWindow: kernel for estimating interference-free frames

    % Convert uint16 frames from camera into double (or float in case of memory overflows)
    frames = im2double(frames);
    
    halfTemporalWindow = temporalWindow/2;

    % Blur `temporalWindow' frames along the positions to estimate per-position
    % interference-free images
    interferenceFree = convn(frames, ones(1, 1, temporalWindow)/temporalWindow, 'same');

    % Pad the interference-free images to remove convolution edge artifacts
    interferenceFree(:, :, 1:halfWindow) = ...
                    repmat(interferenceFree(:, :, halfWindow+1), [1 1 halfWindow]);
    interferenceFree(:, :, end-(halfWindow-1):end) = ...
                    repmat(interferenceFree(:, :, end-halfWindow), [1 1 halfWindow]);

    % Subtract interference-free images to get interference-only images
    interference = (frames-interferenceFree).^2;

    % Blur with a kernel `spatialWindow' in size
    interference = sqrt(imgaussfilt(interference, spatialWindow));

    % Get maximums of interference as ballistic light path image, and positions at maxima
    % as depth
    [direct, idxs] = max(interference, [], 3);
    depth = positions(idxs);

end
\end{lstlisting}
    \caption{Matlab code for acquiring measurements}
    \label{fig:code-process}
\end{figure*}

    {\small
    \bibliographystyle{ieee_fullname}
    \bibliography{passive}}

\end{document}